\documentclass[final]{cvpr}

\usepackage{times}
\usepackage{epsfig}
\usepackage{graphicx}
\usepackage{amsmath}
\usepackage{amssymb}

\usepackage{color}
\usepackage{amsfonts}
\usepackage{multirow}
\usepackage{floatrow}
\usepackage{array}
\newfloatcommand{capbtabbox}{table}[][\FBwidth]

\usepackage{booktabs,tabulary,overpic,xcolor}
\usepackage[caption=false]{subfig}
\definecolor{citecolor}{RGB}{34,139,34}
\usepackage[pagebackref=true,breaklinks=true,letterpaper=true,colorlinks,
  citecolor=citecolor,bookmarks=false]{hyperref}

\usepackage[pagebackref=true,breaklinks=true,colorlinks,bookmarks=false]{hyperref}

\newcommand{\app}{\raise.17ex\hbox{$\scriptstyle\sim$}}

\newcolumntype{x}[1]{>{\centering\arraybackslash}p{#1pt}}

\newlength\savewidth
\newcommand{\tablestyle}[2]{\setlength{\tabcolsep}{#1}\renewcommand{\arraystretch}{#2}\centering\footnotesize}

\makeatletter
\newcommand{\printfnsymbol}[1]{%
  \textsuperscript{\@fnsymbol{#1}}%
}
\makeatother

\def\cvprPaperID{1631} % *** Enter the CVPR Paper ID here
\def\confYear{CVPR 2021}

\begin{document}

%%%%%%%%% TITLE
\title{Collaborative Spatial-Temporal Modeling for Language-Queried\\ Video Actor Segmentation}

\author{Tianrui Hui\textsuperscript{\rm 1,2}\thanks{Equal contribution} \quad Shaofei Huang\textsuperscript{\rm 1,2,5}\printfnsymbol{1} \quad Si Liu\textsuperscript{\rm 3}\thanks{Corresponding author} \quad Zihan Ding\textsuperscript{\rm 3} \quad Guanbin Li\textsuperscript{\rm 4,7} \\ \quad Wenguan Wang\textsuperscript{\rm 6} \quad Jizhong Han\textsuperscript{\rm 1,2} \quad Fei Wang\textsuperscript{\rm 5}\\
\textsuperscript{\rm 1} Institute of Information Engineering, Chinese Academy of Sciences\\
\textsuperscript{\rm 2} School of Cyber Security, University of Chinese Academy of Sciences\\
\textsuperscript{\rm 3} Institute of Artificial Intelligence, Beihang University \\
\textsuperscript{\rm 4} School of Computer Science and Engineering, Sun Yat-sen University \\
\textsuperscript{\rm 5} SenseTime Research 
\quad\textsuperscript{\rm 6} Computer Vision Lab, ETH Zurich
\quad\textsuperscript{\rm 7} Pazhou Lab, Guangzhou \\
}

\maketitle
% \thispagestyle{empty}
% \pagestyle{empty}

%%%%%%%%% ABSTRACT
\begin{abstract}
   Language-queried video actor segmentation aims to predict the pixel-level mask of the actor which performs the actions described by a natural language query in the target frames. 
   Existing methods adopt 3D CNNs over the video clip as a general encoder to extract a mixed spatio-temporal feature for the target frame. 
   Though 3D convolutions are amenable to recognizing which actor is performing the queried actions, it also inevitably introduces misaligned spatial information from adjacent frames, which confuses features of the target frame and yields inaccurate segmentation. 
   Therefore, we propose a collaborative spatial-temporal encoder-decoder framework which contains a 3D temporal encoder over the video clip to recognize the queried actions, and a 2D spatial encoder over the target frame to accurately segment the queried actors. 
   In the decoder, a Language-Guided Feature Selection (LGFS) module is proposed to flexibly integrate spatial and temporal features from the two encoders. 
   We also propose a Cross-Modal Adaptive Modulation (CMAM) module to dynamically recombine spatial- and temporal-relevant linguistic features for multimodal feature interaction in each stage of the two encoders. 
   Our method achieves new state-of-the-art performance on two popular benchmarks 
   with less computational overhead than previous approaches.
\end{abstract}

%%%%%%%%% BODY TEXT
\section{Introduction}
Deep models have achieved notable progress in computer vision and other fields~\cite{gao2020adversarialnas, liao2020ppdm, liao2019gps, huang2020ordnet}. 
Language-queried video actor segmentation~\cite{gavrilyuk2018actor} is an emerging task whose goal is to predict pixel-level mask for the actor performing some actions in a video described by a natural language query. 
Different from language-queried video spatial or temporal localization~\cite{yamaguchi2017spatio, chen2019localizing, anne2017localizing, zhang2019man}, this task requires more fine-grained spatial-temporal modeling and visual-linguistic interaction to generate pixel-level prediction, thus is more challenging. 
At the intersection of computer vision and natural language processing~\cite{huang2020referring, hui2020linguistic, yu2020cross, liao2020real, gao2020interactgan, ren2020scene}, this task enjoys a wide range of applications such as language-driven video editing~\cite{li2020mani}, intelligent surveillance video processing~\cite{sreenu2019intelligent} and human-robot interaction~\cite{qi2020reverie}.

\begin{figure}[t]
   \begin{center}
      \includegraphics[width=1.0\linewidth]{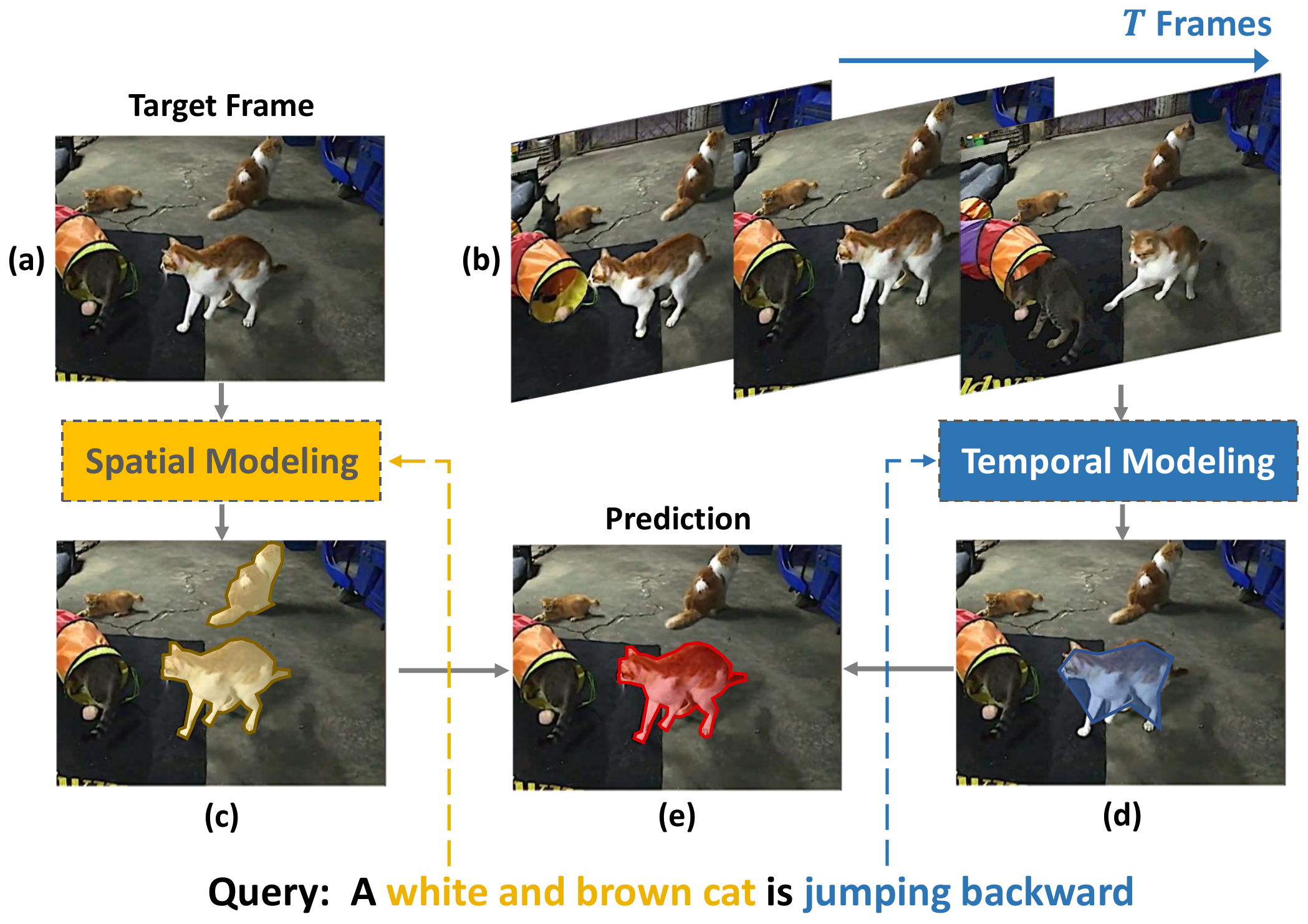}
   \end{center}
      \caption{Illustration of our motivation. (a) The target frame. (b) The input video clip. (c) The spatial encoder can generate fine segmentation but may misidentify other actors due to weak action recognition ability. (d) The temporal encoder can recognize which actor is performing the queried action but may introduce misaligned spatial feature into the target frame, yielding inaccurate segmentation. (e) By integrating spatial and temporal encoders, the correct actor in the target frame can be well segmented.}
   \label{fig:intro}
\end{figure}

As illustrated in Figure~\ref{fig:intro}, given an input query ``a white and brown cat is jumping backward'' and an input video clip (we show $3$ frames for brevity where the target frame is in the middle), language-queried video actor segmentation aims to segment the queried cat on the target frame. 
Since the output is based on the context of the whole video clip, we claim that both temporal modeling over the video clip and spatial modeling over the target frame are essential to solve this task. 
On one hand, as there are two white and brown cats in the target frame, spatial modeling cannot identify the correct cat by exploiting only appearance information. 
It instead inclines to producing fine but false-positive predictions on other cats. 
Therefore, the queried action needs to be recognized by incorporating information from adjacent frames to distinguish the jumping cat from the sitting one, leading to the necessity of temporal modeling over the video clip. 
On the other hand, the jumping cat has various poses and locations in $3$ frames. 
Features of these spatially-misaligned pixels from adjacent frames will disturb the feature representation of the target frame during temporal modeling. 
The correspondence between the feature of the target frame and its ground-truth mask is hence broken. 
Thus, spatial modeling over the target frame is also necessary to provide precise spatial feature.

However, existing approaches~\cite{gavrilyuk2018actor, wang2019asymmetric, mcintosh2020visual, wang2020context,ning2020polar} conduct only temporal modeling over the video clip. 
Concretely, they first feed the video clip into a temporal encoder (3D CNN) to extract cross-frame video features, then apply temporal pooling over the time dimension to obtain a mixed feature of the target frame.
As discussed above, mixing multi-frame spatial information will result in confused spatial feature of the target frame, leading to inaccurate segmentation. 
To tackle this limitation, we propose a collaborative spatial-temporal framework which contains two encoders to conduct spatial modeling over the target frame and temporal modeling over the video clip respectively. 
For the temporal encoder, we adopt a 3D CNN to identify the actor performing the queried action, which can be regarded as the coarse localization of the correct actor by temporal modeling. 
For the spatial encoder, we adopt a 2D CNN to extract precise spatial feature of the target frame, which serves as the fine segmentation of the correct actor by spatial modeling. 
To effectively integrate features from the two encoders, we introduce a Language-Guided Feature Selection (LGFS) module in the decoder to combine the two features with flexible channel selection weights, which are generated from the linguistic feature. 
Thus, language query serves as a selector to form comprehensive spatial-temporal feature for accurate segmentation. 

In addition, language query contains both spatial-relevant information (appearance words, e.g., ``white and brown'') and temporal-relevant information (action words, e.g., ``jumping backward''). 
When interacting with visual feature from the spatial encoder, features of spatial-relevant words should play a more important role than temporal-relevant words and vice versa. 
Therefore, we also propose a Cross-Modal Adaptive Modulation (CMAM) module which dynamically recombines linguistic features by cross-modal attention, yielding spatial- or temporal-relevant linguistic features to adaptively modulate corresponding visual features. 
By densely inserting our CMAM module into each stage of the two encoders, visual features can interact with linguistic features hierarchically and dynamically to highlight regions of the correct actor in spatial and temporal aspects. 

The main contributions of our paper are summarized as follows:
1) We propose a collaborative spatial-temporal framework which contains a temporal encoder to recognize the queried action and a spatial encoder to generate accurate segmentation of the actor. 
A Language-Guided Feature Selection (LGFS) module is proposed in the decoder to aggregate spatial and temporal features comprehensively. 
2) We also propose a Cross-Modal Adaptive Modulation (CMAM) module to conduct spatial- and temporal- relevant multimodal interaction dynamically in each stage of the two encoders. 
3) Extensive experiments on two popular benchmarks show our method outperforms previous state-of-the-art methods with less computational overhead.

\begin{figure*}[t]
   \begin{center}
      \includegraphics[width=0.95\linewidth]{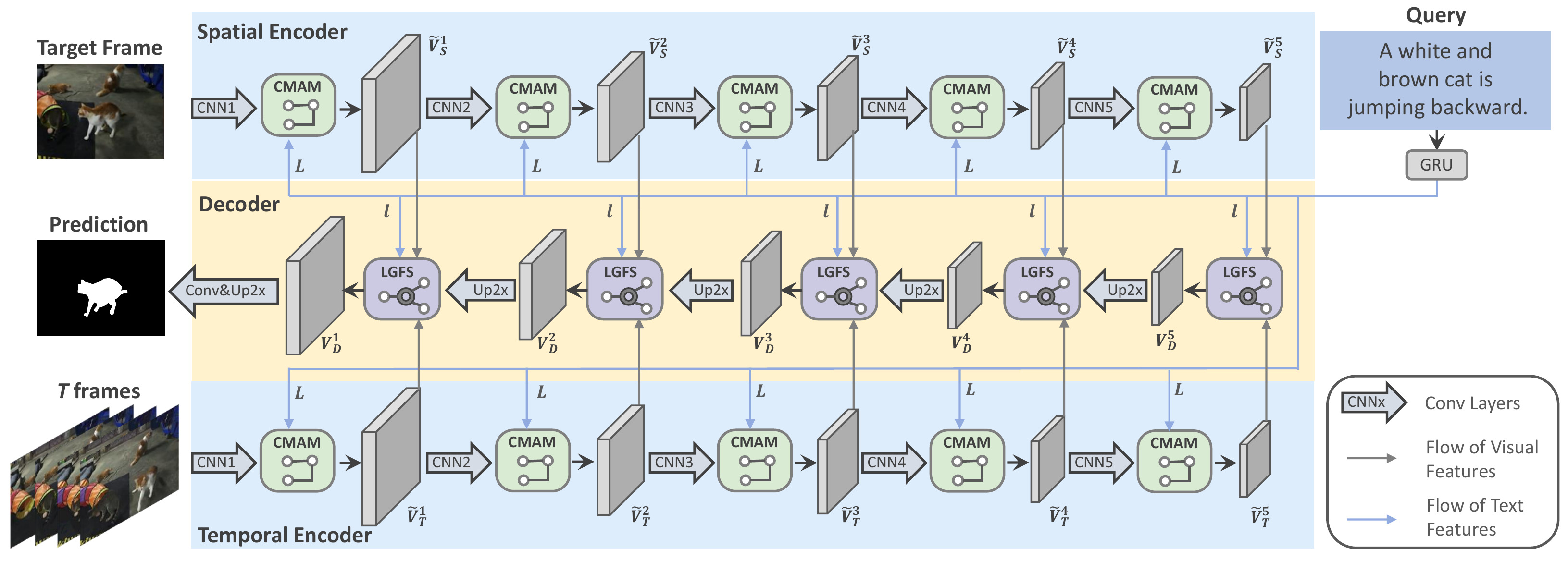}
   \end{center}
      \caption{Overall architecture of our method. Spatial and temporal encoders extract features of the target frame and the video clip respectively, aided by CMAM which dynamically interacts multimodal features in each stage. LGFS is also densely applied in each stage of the decoder to flexibly fuse spatial and temporal features.}
   \label{fig:pipeline}
\end{figure*}

%-------------------------------------------------------------------------
\section{Related Work}
\subsection{Actor and Action Video Segmentation}
To simultaneously infer various types of actors undergoing various actions, Xu~\textit{et al.}~\cite{xu2015can} collect a video dataset (A2D) in which both actors and actions in each video are annotated with pixel-level labels, and introduce a new task named actor and action video segmentation. 
They propose to formulate actors and actions using supervoxels and exploit a trilayer model to reason their relationships for joint labeling. 
Later in~\cite{xu2016actor}, they propose a graphical model to enable long-range interaction modeling among video parts. 
Yan~\textit{et al.}~\cite{yan2017weakly} explore the actor-action segmentation task in a weakly supervised setting with a multi-task ranking model. 
As deep models shown their powerful representation learning ability, Kalogeiton~\textit{et al.}~\cite{kalogeiton2017joint} perform joint actor-action video segmentation via a ``detection-segmentation'' approach. 
Gavrilyuk~\textit{et al.}~\cite{gavrilyuk2018actor} further extend the A2D dataset with natural language descriptions and propose a new task called language-queried video actor segmentation. 
They exploit dynamic filters predicted from language features to convolve with video features and conduct segmentation based on the convolved heatmaps. 
Based on~\cite{gavrilyuk2018actor}, Wang~\textit{et al.}~\cite{wang2020context} incorporate deformable convolutions~\cite{dai2017deformable} into dynamic filters to capture geometric variations. 
ACGA~\cite{wang2019asymmetric} explores co-attention mechanism between video and language features to extract multimodal context for feature enhancement. 
Capsule networks~\cite{sabour2017dynamic} are exploited in~\cite{mcintosh2020visual} to encode video and language features for more effective representations that convolutions. 
PRPE~\cite{ning2020polar} proposes a polar positional encoding method to better localize the actor queried in the video. 
Different from the above works which only use 3D CNNs to extract mixed spatio-temporal feature, we introduce a 2D spatial encoder to collaborate with 3D temporal encoder for compensating the spatial misalignment brought by the temporal encoder. 

\subsection{Language-Queried Video Actor Localization}
Some works have explored the alignment between visual and linguistic modalities by localizing actors and actions in the video by bounding boxes with language queries. 
For 1D temporal localization, Chen~\textit{et al.}~\cite{chen2019localizing} propose a cross-gated attended recurrent network to match video sequence and sentence, and perform cross-frame matching with a self-interactor. 
Local and global video feature integration is explored in~\cite{anne2017localizing} to match with input sentence more comprehensively. 
For 2D spatial localization, Yamaguchi~\textit{et al.}~\cite{yamaguchi2017spatio} first detect persons with spatio-temporal tubes and then conduct matching between tubes and textual descriptions. 
For 3D spatial-temporal localization, Chen~\textit{et al.}~\cite{chen2019weakly} propose a interesting task of localizing a spatial-temporal tube in the video corresponding to the given sentence in a weakly- supervised manner. 
Different from the above works, we identify the actors and actions more precisely with segmentation masks, providing fine-grained multimodal understanding.

\subsection{Spatio-Temporal Modeling}
Spatio-temporal modeling~\cite{simonyan2014two, feichtenhofer2016convolutional, wang2016temporal} is the key to solve video-related tasks. 
A direct way of spatio-temporal modeling is to use 3D CNNs such as C3D~\cite{tran2015learning} and I3D~\cite{carreira2017quo}, etc.
To reduce the computational budget of 3D convolutions, (2+1)D ConvNet~\cite{tran2018closer, qiu2017learning} is proposed to decompose 3D convolution. 
SlowFast~\cite{feichtenhofer2019slowfast} proposes a slow path and a fast path to model spatial and motion information respectively. 
TSM~\cite{lin2019tsm} proposes a temporal shift module which shifts a portion of feature channels along the time dimension, and this operation can be regarded as a special case of 1D temporal convolution. 
In this paper, our model shares the same spirit with SlowFast where a spatial encoder and a temporal encoder are combined to collaboratively extract finer spatio-temporal context for pixel-level classification.

\section{Method}
%-------------------------------------------------------------------------
The overall architecture of our method is illustrated in Figure~\ref{fig:pipeline}. 
The input video clip and query are processed by visual and linguistic encoders respectively (i.e., CNNs~\cite{szegedy2016rethinking} for video data and GRU~\cite{cho2014learning} for sentence data). 
For visual modality, a spatial encoder and a temporal encoder are designed to extract spatial- and temporal-aware visual features respectively. 
In each stage of spatial and temporal visual encoders, visual features and linguistic features are fed into our proposed Cross-Modal Adaptive Modulation (CMAM) module to dynamically highlight visual features matched with the linguistic features. 
Then in the decoder, we propose a Language-Guided Feature Selection (LGFS) module to selectively fuse features of the spatial and temporal encoders from each stage. 
By progressive fusion and upsampling, our decoder produces a feature map of the same size as the input target frame to predict the segmentation mask.

\subsection{Visual and Linguistic Encoders}
Given a video clip with $T$ frames where the target frame annotated with ground-truth mask is in the middle, we adopt Inception V3~\cite{szegedy2016rethinking} as the spatial encoder to process the target frame, and I3D~\cite{carreira2017quo} as the temporal encoder to process the whole video clip. 
We denote features of the $i$-th stage ($i \in [1, 5]$) from the spatial and temporal visual encoders as $V_S^i \in \mathbb{R}^{H^i \times W^i \times C_V^i}$ and $V_T^i \in \mathbb{R}^{T^i \times H^i \times W^i \times C_V^i}$ respectively, where $T^i$, $H^i$, $W^i$, and $C_V^i$ are the frame number, height, width and channel number of the $i$-th visual feature. 
We also adopt an $8$-dimensional coordinate feature to encode relative position information of each pixel following~\cite{wang2019asymmetric}. 
Since the coordinate feature is densely fused with visual features in each stage of the encoders, we omit its denotation in the following formulas for ease of presentation.
For the input textual query with $N$ words, we utilize GRU~\cite{cho2014learning} to extract the linguistic feature which is denoted as $L \in \mathbb{R}^{N \times C_L}$ where $C_L$ denotes the channel number.

\subsection{Cross-Modal Adaptive Modulation}
\begin{figure}[!htbp]
   \begin{center}
      \includegraphics[width=1.0\linewidth]{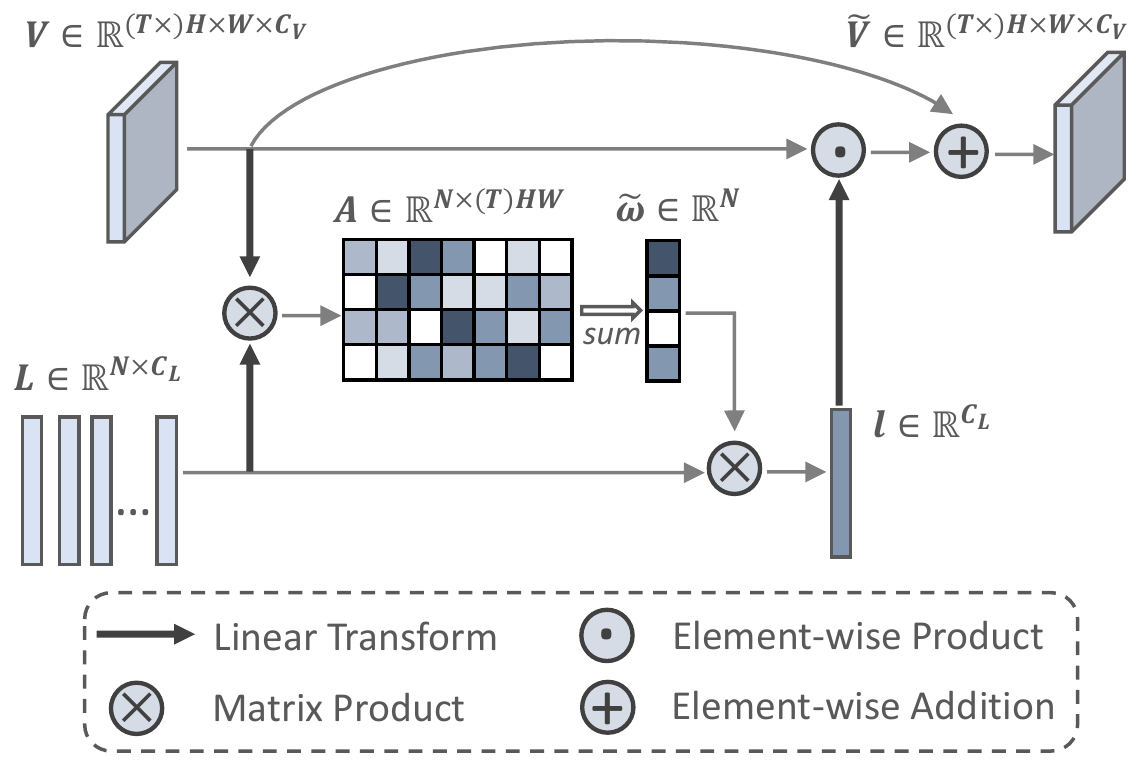}
   \end{center}
      \caption{Illustration of CMAM module. 
      Linguistic feature is dynamically recombined based on the relevance with visual feature. 
      ``S'' and ``T'' subscripts in notations are omitted to denote general spatial or temporal features.}
   \label{fig:cmam}
\end{figure}

Our CMAM aims to enable visual and linguistic features to interact with each other for highlighting visual features which are matched with the corresponding linguistic clues. 
We insert the proposed CMAM module into each stage of the spatial and temporal encoders. 
To clearly elaborate the multimodal interaction process in CMAM, we take the $i$-th stage of our spatial encoder as an example and omit the superscript $i$ for simplicity. 
As illustrated in Figure~\ref{fig:cmam}, given the visual feature $V_S \in \mathbb{R}^{H \times W \times C_V}$ of the target frame and the linguistic feature $L \in \mathbb{R}^{N \times C_L}$ of the sentence, we first conduct cross-modal attention between $V_S$ and $L$ to compute an attention map $A \in \mathbb{R}^{N \times HW}$ which measures the feature relevance between each word and the target frame. 
Concretely, $V_S$ and $L$ are first transformed to the same subspace by convolutions:
\begin{equation}
   \label{eq:trans_v}
   V^\prime_S = Conv_{2d}(V_S),
\end{equation}
\begin{equation}
   \label{eq:trans_l}
   L^\prime_S = Conv_{1d}(L),
\end{equation}
where $V^\prime_S \in \mathbb{R}^{H \times W \times C_M}$, $L^\prime_S \in \mathbb{R}^{N \times C_M}$. 
Then, $V^\prime_S$ is reshaped to $\mathbb{R}^{HW \times C_M}$ to match the matrix dimensions. 
We further perform matrix product between $V^\prime_S$ and $L^\prime_S$ to obtain attention map $A$ as follows:
\begin{equation}
   \label{eq:matmul}
   A = L^\prime_S \otimes {V^\prime_S}^T
\end{equation}
where $\otimes$ denotes matrix product.

Here $A \in \mathbb{R}^{N \times HW}$ measures the relevance between each word and each spatial location. Then we add all the values on the $HW$ dimension and normalize it as follows:
\begin{equation}
   \label{eq:add}
   \begin{split}
   \omega &= \sum_{j = 1}^{HW}A^j, \\
   \tilde{\omega} &= Softmax(\frac{\omega}{\left\lVert \omega \right\rVert_{2}}),
    \end{split}
\end{equation}
where $\left\lVert \cdot \right\rVert_{2}$ denotes the $L_2$ norm of a vector, $A^j \in \mathbb{R}^{N}$ is the feature relevance between the $j$-th spatial location and $N$ words, and $\tilde{\omega} \in \mathbb{R}^{N}$ is the normalized global feature relevance between each word and the whole target frame. 
Therefore, we can use these $N$ weights to linearly re-combine features of $N$ words to attain adaptive sentence feature $l_S = \sum_{k=1}^{N}(\tilde{\omega}^k{L^k}) \in \mathbb{R}^{C_L}$ which contains more spatial information matched with the spatial feature $V_S$ of the target frame. 

Afterwards, a linear layer and $sigmoid$ function are adopted to transform $l_S$ to $\mathbb{R}^{C_V}$ dimensions and generate channel-wise modulation weights $\tilde{l}_S \in \mathbb{R}^{C_V}$:
\begin{equation}
   \label{eq:fc}
   \tilde{l}_S = \sigma(Linear(l_S)),
\end{equation}
where $\sigma$ denotes $sigmoid$ function. 
Inspired by SENet~\cite{hu2018squeeze}, we multiply $\tilde{l}_S$ with feature of the target frame $V_S$ to highlight sentence-relevant visual feature channels and add the modulated feature with original $V_S$ to ease optimization:
\begin{equation}
   \label{eq:senet}
   \tilde{V}_S = V_S + V_S \odot \tilde{l}_S,
\end{equation}
where $\odot$ denotes elementwise product, $\tilde{V}_S$ is the output of the CMAM module and serves as the input feature of the next stage in the spatial visual encoder. 
For the temporal visual encoder, the same operations are applied on all the $T$ frames to highlight sentence-relevant temporal visual features in each stage. 

\subsection{Language-Guided Feature Selection}
In this section, we will elaborate our spatio-temporal decoder which progressively incorporates modulated features from spatial and temporal encoders to recover the feature resolution to the size of the input frame for mask prediction. 
The key problem is which encoder we should trust more for prediction during feature incorporation. 
To solve this problem, we propose a Language-Guided Feature Selection (LGFS) module to select the ratio of spatial and temporal features in the incorporation process under the guidance of linguistic feature. 
Our decoder contains $5$ stages for consistency with the encoders and we take the $i$-th stage ($i \in [1, 5]$) as an example to detail our LGFS module. 

Concretely, we conduct average pooling on the linguistic features $L \in \mathbb{R}^{N \times C_L}$ to obtain feature of the whole sentence $l \in \mathbb{R}^{C_L}$. 
Since $l$ contains both spatial- and temporal-relevant linguistic information, we can use it to capture channel dependencies between spatial and temporal visual features inspired by SKNet~\cite{li2019selective}, which proves to be effective on image classification. 
Two linear layers are applied on $l$ to generate raw selection weights of each pair of channels from spatial and temporal visual features as follows:
\begin{equation}
   \label{eq:add_lang_s}
   g_S = Linear(l),~~g_T = Linear(l),
\end{equation}
where $g_S \in \mathbb{R}^{C_V}$ and $g_T \in \mathbb{R}^{C_V}$ have the same channel number with visual features. 
We apply $Softmax$ over each pair of channels of $g_S$ and $g_T$ to produce the normalized channel selection weights $\tilde{g}_S$ and $\tilde{g}_T$. 
The incorporated feature $V_F^i$ is obtained as follows:
\begin{equation}
   \label{eq:sknet}
   V_F^i = \tilde{V}_S^{i} \odot \tilde{g}_S + \tilde{V}_T^{i} \odot \tilde{g}_T,
\end{equation}
where $\odot$ denotes elementwise product using broadcasting rule. We slightly abuse the notation of $\tilde{V}_T^{i}$ to denote the modulated feature of the target frame from the temporal encoder for simplicity. 
Finally, the output feature of the $i$-th stage in our decoder $V_D^i \in \mathbb{R}^{H^i \times W^i \times C_V^i}$ is defined as:
\begin{equation}
   V_D^i = 
   \begin{cases}
      V_F^i, & i = 5, \\
      V_F^i + Upsample(V_D^{i+1}), & 1 \leq i \leq 4.
   \end{cases}
\end{equation}

\begin{table*}[t]
   \centering
   \begin{tabular}{|m{4.5cm}<{\centering}|c|c|c|c|c|c|c|c|}
         \hline
         \multirow{2}*{\textbf{Method}} & \multicolumn{5}{c|}{\textbf{Precision}} & \textbf{AP} & \multicolumn{2}{c|}{\textbf{IoU}} \\
         \cline{2-9}
         & P@0.5 & P@0.6 & P@0.7 & P@0.8 & P@0.9 & 0.5:0.95 & Overall & Mean \\
         \hline
         Hu \textit{et al.}~\cite{hu2016segmentation}~\scriptsize{ECCV2016} & 34.8 & 23.6 & 13.3 & 3.3 & 0.1 & 13.2 & 47.4 & 35.0 \\
         Li \textit{et al.}~\cite{li2017tracking}~\scriptsize{CVPR2017} & 38.7 & 29.0 & 17.5 & 6.6 & 0.1 & 16.3 & 51.5 & 35.4 \\
         Gavrilyuk \textit{et al.}~\cite{gavrilyuk2018actor}~\scriptsize{CVPR2018} & 47.5 & 34.7 & 21.1 & 8.0 & 0.2 & 19.8 & 53.6 & 42.1 \\
         Gavrilyuk \textit{et al.}~\cite{gavrilyuk2018actor}$\dagger$~\scriptsize{CVPR2018} & 50.0 & 37.6 & 23.1 & 9.4 & 0.4 & 21.5 & 55.1 & 42.6 \\
         ACGA~\cite{wang2019asymmetric}~\scriptsize{ICCV2019} & 55.7 & 45.9 & 31.9 & 16.0 & 2.0 & 27.4 & 60.1 & 49.0 \\
         VT-Capsule~\cite{mcintosh2020visual}~\scriptsize{CVPR2020} & 52.6 & 45.0 & 34.5 & 20.7 & 3.6 & 30.3 & 56.8 & 46.0 \\
         CMDy~\cite{wang2020context}~\scriptsize{AAAI2020} & 60.7 & 52.5 & 40.5 & 23.5 & 4.5 & 33.3 & 62.3 & 53.1 \\
         PRPE~\cite{ning2020polar}~\scriptsize{IJCAI2020} & 63.4 & 57.9 & 48.3 & 32.2 & 8.3 & 38.8 & 66.1 & 52.9 \\
         \hline
         Ours & \textbf{65.4} & \textbf{58.9} & \textbf{49.7} & \textbf{33.3} & \textbf{9.1} & \textbf{39.9} & \textbf{66.2} & \textbf{56.1} \\
         \hline
   \end{tabular}
   \caption{Comparison with state-of-the-art methods on A2D Sentences test set. Our method significantly outperforms previous methods using only RGB input. $\dagger$ denotes utilizing additional optical flow input.}
   \label{tab:sota_a2d}
\end{table*}

\begin{table*}[!htbp]
   \centering
   \begin{tabular}{|m{4.5cm}<{\centering}|c|c|c|c|c|c|c|c|}
         \hline
         \multirow{2}*{\textbf{Method}} & \multicolumn{5}{c|}{\textbf{Precision}} & \textbf{AP} & \multicolumn{2}{c|}{\textbf{IoU}} \\
         \cline{2-9}
         & P@0.5 & P@0.6 & P@0.7 & P@0.8 & P@0.9 & 0.5:0.95 & Overall & Mean \\
         \hline
         Hu \textit{et al.}~\cite{hu2016segmentation}~\scriptsize{ECCV2016} & 63.3 & 35.0 & 8.5 & 0.2 & 0.0 & 17.8 & 54.6 & 52.8 \\
         Li \textit{et al.}~\cite{li2017tracking}~\scriptsize{CVPR2017} & 57.8 & 33.5 & 10.3 & 0.6 & 0.0 & 17.3 & 52.9 & 49.1 \\
         Gavrilyuk \textit{et al.}~\cite{gavrilyuk2018actor}~\scriptsize{CVPR2018} & 69.9 & 46.0 & 17.3 & 1.4 & 0.0 & 23.3 & 54.1 & 54.2 \\
         Gavrilyuk \textit{et al.}~\cite{gavrilyuk2018actor}$\ddagger$~\scriptsize{CVPR2018} & 71.2 & 51.8 & 26.4 & 3.0 & 0.0 & 26.7 & 55.5 & 57.0 \\
         ACGA~\cite{wang2019asymmetric}~\scriptsize{ICCV2019} & 75.6 & 56.4 & 28.7 & 3.4 & 0.0 & 28.9 & 57.6 & 58.4 \\
         VT-Capsule~\cite{mcintosh2020visual}~\scriptsize{CVPR2020} & 67.7 & 51.3 & 28.3 & 5.1 & 0.0 & 26.1 & 53.5 & 55.0 \\
         CMDy~\cite{wang2020context}~\scriptsize{AAAI2020} & 74.2 & 58.7 & 31.6 & 4.7 & 0.0 & 30.1 & 55.4 & 57.6 \\
         PRPE~\cite{ning2020polar}~\scriptsize{IJCAI2020} & 69.0 & 57.2 & 31.9 & 6.0 & \textbf{0.1} & 29.4 & - & - \\
         \hline
         Ours & \textbf{78.3} & \textbf{63.9} & \textbf{37.8} & \textbf{7.6} & 0.0 & \textbf{33.5} & \textbf{59.8} & \textbf{60.4} \\
         \hline
   \end{tabular}
   \caption{Comparison with state-of-the-art methods on J-HMDB Sentences test set using the best model trained on A2D Sentences \textbf{without finetuning}. Our method shows notable generalization ability. $\ddagger$ denotes training more layers of I3D backbone on A2D Sentences.}
   \label{tab:sota_jhmdb}
\end{table*}

%-------------------------------------------------------------------------
\section{Experiments}
\subsection{Datasets and Evaluation Metrics}
We conduct experiments on two popular language-queried video actor segmentation benchmarks including A2D Sentences~\cite{gavrilyuk2018actor} and J-HMDB Sentences~\cite{gavrilyuk2018actor}. 
We adopt Overall IoU, Mean IoU and Precision@$X$ (P@$X$) as metrics to evaluate our model following prior works~\cite{wang2019asymmetric, wang2020context}. 
Overall IoU calculates the ratio of the accumulated intersection area over the accumulated union area between predictions and ground-truth masks on all the test samples, while Mean IoU calculates the averaged IoU over all the test samples. 
Precision@$X$ measures the percentage of test samples whose IoU are higher than a predefined threshold $X$, where $X \in [0.5, 0.6, 0.7, 0.8, 0.9]$. 
We also compute the Average Precision (AP) over the section of $[0.50:0.05:0.95]$.

\subsection{Implementation Details}
Following prior works~\cite{wang2019asymmetric, wang2020context, ning2020polar}, we use the I3D~\cite{carreira2017quo} networks pretrained on Kinetics$400$~\cite{carreira2017quo} dataset as our temporal visual encoder. 
For spatial visual encoder, we adopt Inception V3~\cite{szegedy2016rethinking} pretrained on ImageNet~\cite{deng2009imagenet} dataset. 
We adopt two GRUs for the two visual encoders to extract linguistic features respectively. 
The maximum length of the input sentence is set as $20$. 
We sample $T = 8$ RGB frames as the video input to our model where the annotated target frame is in the middle. 
The input frames are resized and padded to $320 \times 320$. 
Adam~\cite{kingma2014adam} is utilized as the optimizer and the training process is divided into two stages. 
First, we train the spatial and temporal networks (both encoders and decoders) respectively on A2D Sentence dataset for $12$ epochs with batch size $8$ and learning rate $5e^{-4}$ (divided by $10$ every $8$ epochs). 
Then, the two pretrained encoders are combined with a random-initialized decoder and fixed during finetuning the decoder for another $4$ epochs with the same learning rate $5e^{-4}$.

\subsection{Comparison with State-of-the-arts}
We conduct experiments on A2D Sentences and J-HMDB Sentences to compare our method with pervious state-of-the-arts. 
As illustrated in Table~\ref{tab:sota_a2d}, our method outperforms pervious state-of-the-arts on A2D Sentences test set, indicating the effectiveness of collaborative learning of spatial and temporal encoders and adaptive visual-linguistic interaction. 
Comparing with CMDy~\cite{wang2020context} and PRPE~\cite{ning2020polar}, our method achieves $3.0\%$ and $3.2\%$ absolute improvements on Mean IoU respectively. 
For the most rigorous metric P@0.9, our method is also superior than the previous performances of CMDy and PRPE, demonstrating that our method can not only accurately identify the correct actor through cross-modal alignment, but also generate complete mask to cover the actor. 
Since Overall IoU favors large actors while Mean IoU treating actors of different scales equally, our improvements on IoU metrics also show that our method can well handle the scale variation of actors.

We further verify the generalization ability of our method on J-HMDB Sentences test set. 
Following prior works~\cite{ning2020polar, wang2020context, wang2019asymmetric}, we use the best model pretrained on A2D Sentences dataset to directly evaluate all the test samples in J-HMDB Sentences without finetuning. 
For each testing video, $3$ frames are uniformly sampled to evaluate the performance. 
As shown in the Table~\ref{tab:sota_jhmdb}, our method accomplishes significant performance gains over previous state-of-the-arts, indicating that our method can excavate richer multimodal information through the mutual enhancement of spatial and temporal encoders, so as to obtain stronger generalization ability. 
Noted that all the methods including ours produce approximate zero performance on P@0.9, which is probably because models without training on J-HMDB Sentences cannot generate particularly fine masks on unseen samples.

\begin{table*}[t]
   % % subfloat a - spatial temporal encoders
   \subfloat[\textbf{Component analysis.} Verifying the effectiveness of each component in our encoder-decoder framework. ``Spatial'' and ``Temporal'' denote spatial and temporal encoders respectively.
   \label{tab:ablation:encoders}]{
   \tablestyle{6pt}{1.15}\begin{tabular}{|c|c|c|c|c|c|c|c|c|c|c|c|c|}
      \hline
      & \multirow{2}*{Spatial} & \multirow{2}*{Temporal} & \multirow{2}*{LGFS} & \multirow{2}*{CMAM} & \multicolumn{5}{c|}{\textbf{Precision}} & \textbf{AP} & \multicolumn{2}{c|}{\textbf{IoU}} \\
      \cline{6-13}
      & & & & & P@0.5 & P@0.6 & P@0.7 & P@0.8 & P@0.9 & 0.5:0.95 & Overall & Mean \\
      \hline
      1 & $\surd$ & & & & 53.0 & 45.7 & 35.1 & 20.3 & 4.4 & 29.0 & 56.7 & 47.7 \\
      2 & & $\surd$ & & & 54.4 & 45.5 & 33.7 & 18.1 & 2.9 & 28.1 & 58.2 & 48.2 \\
      3 & $\surd$ & $\surd$ & & & 58.5 & 51.8 & 42.3 & 27.9 & 7.5 & 34.5 & 62.2 & 51.7 \\
      4 & $\surd$ & $\surd$ & $\surd$ & & 60.3 & 53.6 & 44.0 & 29.1 & 7.9 & 36.0 & 62.8 & 52.9 \\
      5 & $\surd$ & $\surd$ & $\surd$ & $\surd$ & \textbf{65.4} & \textbf{58.9} & \textbf{49.7} & \textbf{33.3} & \textbf{9.1} & \textbf{39.9} & \textbf{66.2} & \textbf{56.1} \\
      \hline
   \end{tabular}}

   % subfloat b - LGFS
   \subfloat[\textbf{Spaital and temporal feature fusion.} % Exploiting language as guidance can obtain higher performance.
   \label{tab:ablation:lgfs}]{
   \tablestyle{3.5pt}{1.15}\begin{tabular}{|c|c|c|c|c|c|c|c|c|}
      \hline
      \multirow{2}*{\textbf{ST-Fusion}} & \textbf{AP} & \multicolumn{2}{c|}{\textbf{IoU}} \\
      \cline{2-4}
      & 0.5:0.95 & Overall & Mean \\
      \hline
      Add & 34.5 & 62.2 & 51.7 \\
      Max & 34.8 & 62.6 & 51.6 \\
      LGFS (Ours) & \textbf{36.0} & \textbf{62.8} & \textbf{52.9} \\
      \hline
   \end{tabular}}\hspace{3mm}
   % subfloat c - CMAM
   \subfloat[\textbf{Inserting positions of CMAM.} $I_j$ denotes the $j$-th stage in Inception and I3D.
   \label{tab:ablation:cmam}]{
   \tablestyle{3.5pt}{1.15}\begin{tabular}{|c|c|c|c|c|c|c|c|c|}
      \hline
      \multirow{2}*{\textbf{Position}} & \textbf{AP} & \multicolumn{2}{c|}{\textbf{IoU}} \\
      \cline{2-4}
      & 0.5:0.95 & Overall & Mean \\
      \hline
      $\{I_5\}$ & 36.9 & 63.1 & 53.3 \\
      $\{I_5, I_4\}$ & 38.5 & 65.2 & 55.0 \\
      $\{I_5, I_4, I_3\}$ & 39.0 & 65.4 & 55.6 \\
      $\{I_5, I_4, I_3, I_2\}$ & 39.4 & 65.9 & 55.7 \\
      $\{I_5, I_4, I_3, I_2, I_1\}$ & \textbf{39.9} & \textbf{66.2} & \textbf{56.1} \\
      \hline
   \end{tabular}}\hspace{3mm}
   % subfloat d - backbones
   \subfloat[\textbf{Backbones for our spatial and temporal encoder.} R50: ResNet-50. IV3: Inception V3.
   \label{tab:ablation:backbones}]{
   \tablestyle{3.5pt}{1.15}\begin{tabular}{|c|c|c|c|c|c|c|c|c|}
      \hline
      \multirow{2}*{\textbf{Backbone}} & \textbf{AP} & \multicolumn{2}{c|}{\textbf{IoU}} \\
      \cline{2-4}
      & 0.5:0.95 & Overall & Mean \\
      \hline
      R50 + S3D & 38.5 & \textbf{67.1} & 55.1 \\
      R50~\cite{he2016deep} + I3D & 39.5 & 66.4 & \textbf{56.4} \\
      IV3 + S3D~\cite{xie2018rethinking} & 38.3 & 66.7 & 55.0 \\
      IV3 + I3D & \textbf{39.9} & 66.2 & 56.1 \\
      \hline
   \end{tabular}}
   % main caption
   
   \caption{\textbf{Ablation studies}. Models are trained on A2D Sentences \texttt{train} split and evaluated on \texttt{test} split.}
   \label{tab:ablations}
\end{table*}

\subsection{Ablation Studies}
We conduct ablation studies on the A2D Sentences dataset to evaluate different design of our framework. 

\textbf{Component Analysis.}
We summarize the ablation results of our proposed encoders and modules in Table~\ref{tab:ablation:encoders}.
The $1$st and $2$nd row denote our spatial- and temporal-only baselines where multimodal interactions only occurs in the decoders by visual and linguistic feature concatenation. 
Spatial and temporal baselines outperform each other on P@0.9 and P@0.5 respectively, which indicates spatial encoder can yield finer segmentation while temporal encoder can identify the actors more accurately. 
When simply combining the two encoders together in the $3$rd row, we can observe an obvious performance boost in all metrics, which well demonstrates the complementarity of spatial and temporal encoders. 
Incorporating our LGFS module is able to further improve the performance, which shows that using language information as guidance can select effective spatial and temporal features more flexibly. 
In the $5$th row, our CMAM module also brings large performance gain over a strong result in the $4$th row. 
We insert CMAM into each stage of the spatial and temporal encoders to conduct multimodal feature interaction and remove the feature concatenation in our baselines. 
Results of CMAM prove the effectiveness of modulating visual features by dynamically recombine spatial- and temporal-relevant linguistic features.

\textbf{Spatial-Temporal Feature Fusion.}
Table~\ref{tab:ablation:lgfs} presents results of different spatial-temporal feature fusion methods without CMAM module. 
Elementwise addition and maximization produce similar results. 
However, these simple fusion methods usually lack the ability to select appropriate spatial and temporal visual information according to the needs of language information. 
Our LGFS compensates for this deficiency and outperforms the above two operations, demonstrating the effectiveness of language guidance.

\textbf{Inserting Positions of CMAM.}
We evaluate different inserting positions of CMAM and summarize the results in Table~\ref{tab:ablation:cmam}. 
Inserting CMAM into the $5$th and $4$th stages of our spatial-temporal encoders can bring relatively significant improvements, which is probably because features from deep layers usually contain more high-level semantic information and are beneficial to the segmentation. 
As we insert CMAM into the shallow layers of encoders, segmentation performance is also constantly improved, showing the dynamically recombined linguistic features can modulate visual features of different abstraction levels. 

\textbf{Backbones.} 
Table~\ref{tab:ablation:backbones} shows the results of different backbone selections for our spatial and temporal encoders. 
We conduct experiments on ResNet-50~\cite{he2016deep} and S3D~\cite{xie2018rethinking}, which have similar capacities with their Inception and I3D counterparts. 
From the table we can observe that models with different backbones exhibit stable and high performances, which demonstrates that our collaborative spatial-temporal framework can well adapt to different backbones. 
For fair comparison with previous approaches, we adopt I3D and Inception V3 (which has moderate computational overhead comparing with ResNet-50) as our temporal and spatial encoders.

\begin{table}[!htbp]
   \centering
   \begin{tabular}{|c|c|c|c|}
      \hline
      \textbf{Method} & \textbf{Input Size} & \textbf{GFLOPs} & \textbf{AP} \\
      \hline
      ACGA~\cite{wang2019asymmetric} & 16 $\times$ 512 $\times$ 512 & 630.83 & 27.4 \\
      CMDy~\cite{wang2020context} $\dagger$ & 16 $\times$ 512 $\times$ 512 & $>$ 600 & 33.3 \\
      Ours & 8 $\times$ 320 $\times$ 320 & 213.06 & \textbf{39.9} \\
      Ours-Spa $\ddagger$ & 8 $\times$ 320 $\times$ 320 & 19.54 & - \\
      \hline
   \end{tabular}
   \caption{Computational overhead analysis. The RGB input size is \emph{Frames} $\times$ \emph{Height} $\times$ \emph{Width} ($3$ channels are omitted). $\dagger$ denotes the GFLOPs is estimated. $\ddagger$ denotes we only calculate the GFLOPs of our spatial encoder part, whose performance is not available.}
   \label{tab:computation}
\end{table}

\subsection{Computational Overhead}
We calculate the computational overhead of previous methods and ours in Table~\ref{tab:computation}. 
Since CMDy has not released code, we estimate its computational overhead according to the details in their paper, including I3D backbone and the same input size as reported in the paper of ACGA. 
Both CMDy and ACGA rely on large input size to retain their performances. 
However, our method outperforms theirs with significant margins using $3\times$ less GFLOPs and much smaller input size, showing that our method is more efficient in excavating useful information from multimodal features. 
Noted that our introduced spatial encoder only takes up $9.2\%$ computational overhead of our full framework but brings considerable performance improvement, which well demonstrates the effectiveness of our collaborative spatial-temporal modeling framework.

\begin{figure*}[t]
   \begin{center}
      \includegraphics[width=0.85\linewidth]{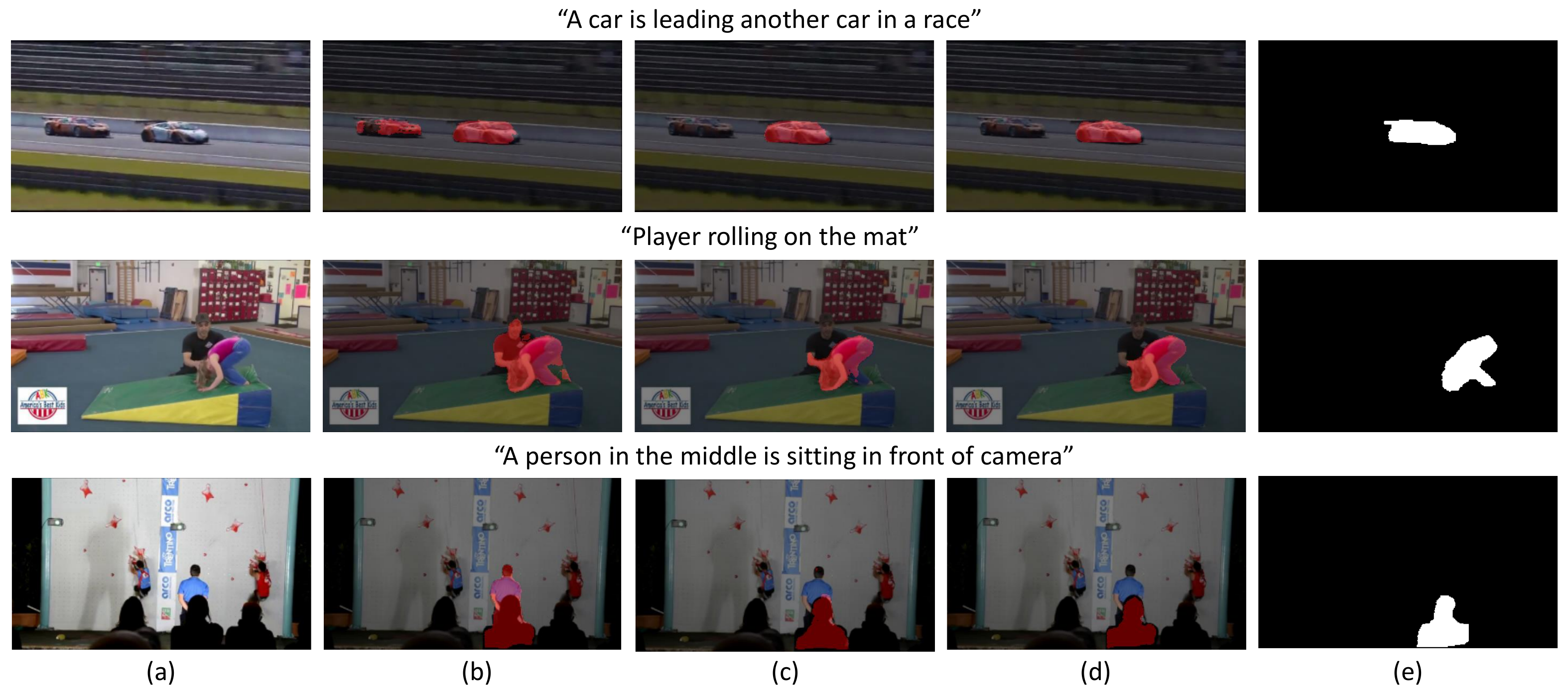}
   \end{center}
      \caption{Qualitative analysis on target frames. (a) Target frame. (b) Results of our model using spatial encoder only. (c) Results of our model using temporal encoder only. (d) Results of our model using spatial and temporal encoders. (e) Ground-truth.}
   \label{fig:visualize}
\end{figure*}

\begin{figure}[!htbp]
   \begin{center}
      \includegraphics[width=0.95\linewidth]{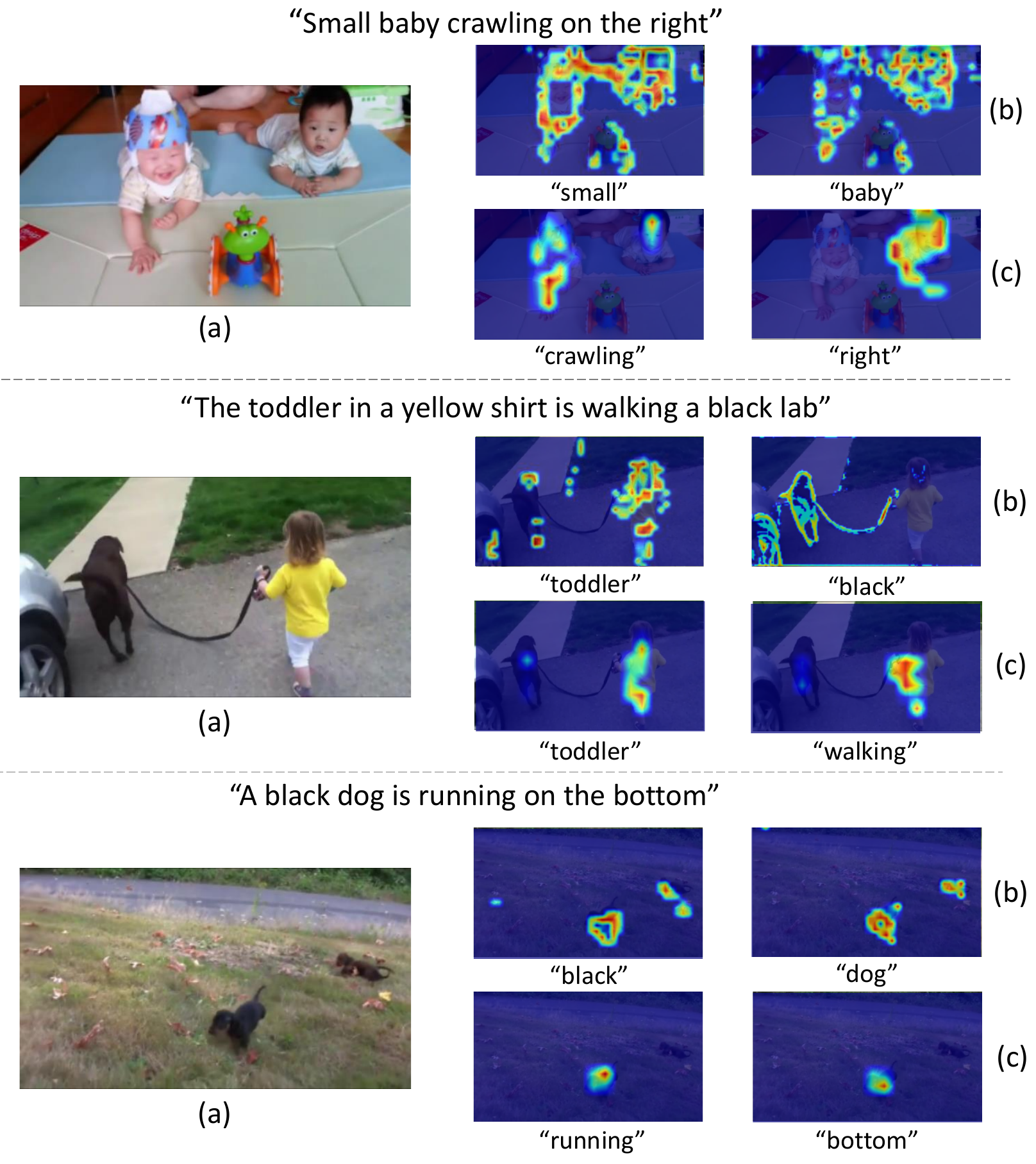}
   \end{center}
      \caption{Visualization of attention maps between words and frames in CMAM. (a) Target frame. (b) Attention maps in spatial encoder. (c) Attention maps in temporal encoder.}
   \label{fig:attn}
\end{figure}

\subsection{Qualitative Analysis}
Figure~\ref{fig:visualize} presents the visualization results of our model on target frames, which provides qualitative analysis on the complementarity of our spatial and temporal encoders. 
As shown in the $2$nd row, using only spatial encoder tends to make false-positive predictions on other actors (e.g., the man) which are irrelevant with the description due to the unawareness of action information, albeit the generated masks are relatively accurate. 
When only temporal encoder is used, the woman who is rolling can be correctly located but the segmentation result lacks some local details. 
For example, part of the woman's legs is misclassified as background. 
Incorporating both the spatial and temporal encoders yields precise segmentation on the correct actor. 
Similar phenomenon also appears in other examples of Figure~\ref{fig:visualize}. 

We also visualize the attention maps between words and target frames in CMAM from spatial and temporal encoders in Figure~\ref{fig:attn}. 
In the $1$st row, spatial-relevant words ``small'' and ``baby'' yield highly-responsive attention maps on body of the two babies in the spatial encoder, while temporal-relevant word ``crawling'' mainly focuses on the moving hands and heads of the two babies in the temporal encoder. 
The word ``running'' in the $3$rd row also focuses on the moving black dog to capture actions in the temporal encoder.
These results show CMAM can well associate visual and linguistic features using spatial and temporal information.

Noted that for actor who only appears in a part of the video, our method can handle such case. 
For each frame, through our spatial and temporal encoders, intra-frame visual information and global cues from the whole video clip are collected and fused together. 
Then, the final segment per frame is guided by linguistic cues. 
Thus, our model does not explicitly leverage any causal information or frame-by-frame propagation.

%-------------------------------------------------------------------------
\section{Conclusion and Future Work}
In this paper, we explore the language-queried video actor segmentation task. 
We propose a collaborative encoder-decoder framework containing a 3D temporal encoder to recognize the queried actions and a 2D spatial encoder to well segment the actors, which alleviates the spatial misalignment issue brought by 3D CNNs in prior works. 
An LGFS module is introduced in the decoder to flexibly fuse spatial-temporal features. 
In addition, we also propose a CMAM module to dynamically recombine linguistic features for more adaptive multimodal feature interaction in each encoder. 
Our method outperforms previous methods by large margins on two popular benchmarks with $3\times$ less computational overhead. 
In the future, we hope to accelerate current framework by designing more lightweight spatial and temporal encoders. 

\vspace{-2mm}
\paragraph{Acknowledgments} This research is supported in part by National Natural Science Foundation of China (Grant 61876177, 61976250 and U1811463), Beijing Natural Science Foundation (4202034), Fundamental Research Funds for the Central Universities, the Guangdong Basic and Applied Basic Research Foundation (No. 2020B1515020048), Zhejiang Lab (No. 2019KD0AB04), SenseTime Group Ltd. and CCF-Baidu Open Fund.

{\small
\bibliographystyle{ieee_fullname}
\bibliography{1631_final_arxiv}
}

\end{document}